# Subspace Selection to Suppress Confounding Source Domain Information in AAM Transfer Learning

Azin Asgarian[1], Ahmed Bilal Ashraf[2], David Fleet[1], and Babak Taati[1,2,3]

[1]Department of Computer Science, University of Toronto
[2]Toronto Rehabilitation Institute, University Health Network
[3]Institute of Biomaterials & Biomedical Engineering, University of Toronto

## Abstract

*Active appearance models (AAMs) are a class of generative models that have seen tremendous success in face analysis. However, model learning depends on the availability of detailed annotation of canonical landmark points. As a result, when accurate AAM fitting is required on a different set of variations (expression, pose, identity), a new dataset is collected and annotated. To overcome the need for time consuming data collection and annotation, transfer learning approaches have received recent attention. The goal is to transfer knowledge from previously available datasets (source) to a new dataset (target). We propose a subspace transfer learning method, in which we select a subspace from the source that best describes the target space. We propose a metric to compute the directional similarity between the source eigenvectors and the target subspace. We show an equivalence between this metric and the variance of target data when projected onto source eigenvectors. Using this equivalence, we select a subset of source principal directions that capture the variance in target data. To define our model, we augment the selected source subspace with the target subspace learned from a handful of target examples. In experiments done on six publicly available datasets, we show that our approach outperforms the state of the art in terms of the RMS fitting error as well as the percentage of test examples for which AAM fitting converges to the ground truth.*

## 1. Introduction

Active appearance models (AAMs) are deformable generative models used to capture shape and appearance variation for various computer vision applications [1]. AAMs have been very successful in applications where the objects of interest have spatial correspondence in structure, *e.g.* face analysis (pre-processing for identity recognition, pose-estimation, emotion recognition) [2–4], hand analysis (hand and gesture recognition, accessibility applications) [5, 6], and 3D brain segmentation [7].

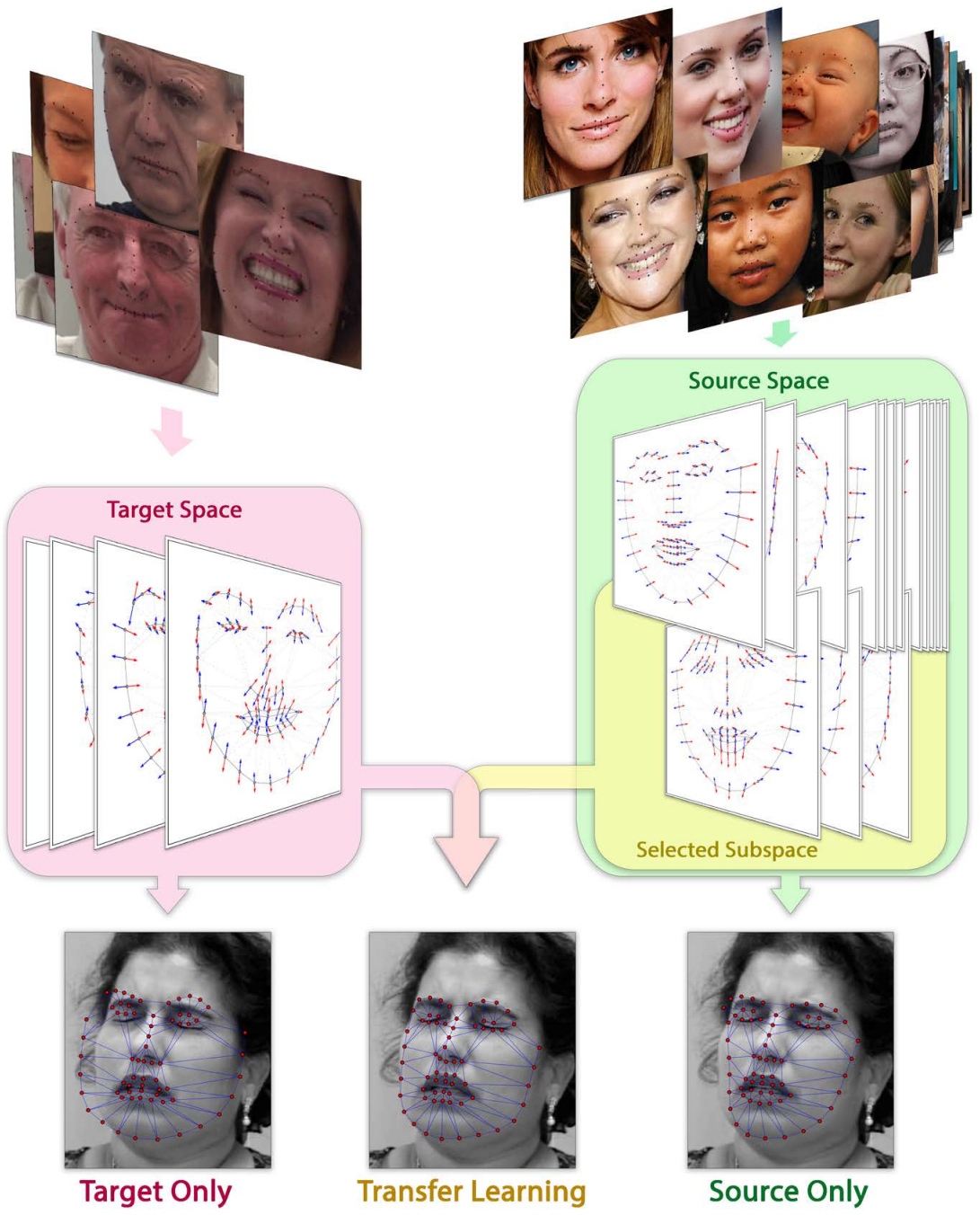

Figure 1. Overview of *Subspace Selection* AAM transfer learning. The left most column shows the case when the model is based only on target domain examples. The right most column shows the case when the model is based on the principal directions of the source domain. In the middle, we show the situation where the target model is augmented with those source principal directions which capture target data variance.

The AAM model is constructed by building a statistical shape model from a set of annotated landmark points which are predefined to describe the shape of an object *e.g.* the face. An appearance model is also built using a set of images that are warped to a canonical reference shape (usually the mean of the shape model). For a given test image, model fitting involves finding the best combination of shape and appearance parameters such that the reconstructed shape can be warped into the reference frame; and the input image (subjected to the same warp) is best described by the reconstructed image based on the appearance parameters.

A key requirement for learning is the availability of a training set consisting of images and detailed annotation of landmark points. In an attempt to capture variations in pose, expression, illumination, and identity, a number of face datasets have been collected and annotated (*e.g.* [8,9]). Although the collection and annotation of datasets was necessary to ensure that models capture enough variation, it has also led the community to realize that collecting more and more data might not be the best approach [10, 11].

Historically, every time AAM fitting was required on a different set of variations, a new dataset would be collected and annotated. Despite this effort over the last two decades, the generalizability of AAMs to new domains remains challenging [10]. To overcome this issue, transfer learning has received attention from the AAM research community [11]. Transfer learning techniques attempt to improve generalizability of AAMs to a new set of data (referred to as the *target* domain) through the transfer of knowledge learned from a pre-existing set of data (referred to as the *source* domain).

In this paper, for transferring the learned knowledge we focus on the principal directions of the source dataset and propose to judiciously select a subspace from the space of source data. We argue that not all principal source directions are useful, and some directions might even be detrimental to the AAM fitting process. We propose a method to weed out the confounding directions that are not representative of target subspace. Specifically, we retain the source basis vectors which co-vary with target examples and capture the variance in the target dataset. As a result, we end up discarding source basis vectors that do not co-vary with the target data. We then combine these retained source directions with the target subspace determined by a handful of target examples. An overview of our approach is shown in Figure 1.

Specific contributions in this paper are as follows:

- We propose a metric to compute the directional similarity between source eigenvectors and the target subspace. We show an equivalence between the directional similarity metric and the variance of target data when projected onto source eigenvectors (§3).
- Capitalizing on this equivalence, we present a method to select a subspace from source shape and appearance models by picking a subset of principal directions that capture the variance in target data. This retained subset encodes the knowledge transferred from the source to the target domain (§3).
- In experiments on six publicly available datasets, we show that our approach outperforms a series of baselines including state of the art AAM transfer learning method [11] (§5). Also we present several insights that emerge by analyzing the source directions selected by our algorithm (§5.1).

## 2. Background

We begin with a brief description of AAMs and their training process, following [12]. We then review existing literature on transfer learning methods for AAMs.

### 2.1. Active Appearance Models: A Review

An AAM is a generative model that captures variation of shape and appearance from a set of labeled examples. The model thus has two parts, one for shape, and another for appearance.

**Shape Model:** A shape $\mathbf{s}$ is represented by a 2D mesh of $V$ vertices, $\mathbf{s} = (x_1, y_1, x_2, y_2, ..., x_V, y_V)^T$. Consider a set of $N$ training samples $\{(\mathbf{I}_i, \mathbf{s}_i) \mid i \in \{1, ..., N\}\}$, each consisting of an image $\mathbf{I}_i$ and its corresponding shape $\mathbf{s}_i$. To build a model, the shape samples $(\mathbf{s}_1, \mathbf{s}_2, ..., \mathbf{s}_N)$ are first aligned using Generalized Procrustes Analysis [12]. The outcome of procrustes analysis is a global 2D similarity transformation which can also be modeled by a linear combination of four basis vectors [12]. The set of four basis vectors modeling the global transformation will be denoted $\mathbf{\Phi}_g$. Let the matrix $\mathbf{Z} = [\mathbf{z}_1, \mathbf{z}_2, ..., \mathbf{z}_N] \in \mathbb{R}^{2V \times N}$ consist of the aligned shape samples. By applying PCA to $\mathbf{Z}$ we get the orthonormal shape eigenvectors (that model local shape variation) $\mathbf{\Phi}_l \in \mathbb{R}^{2V \times K}$ (where $K < N$), and the corresponding eigenvalues $\boldsymbol{\lambda} \in \mathbb{R}^K$. $\mathbf{\Phi}_g$ and $\mathbf{\Phi}_l$ are then combined to give the shape model $\{\boldsymbol{\mu}, \mathbf{\Phi}\}$, where $\mathbf{\Phi} \in \mathbb{R}^{2V \times (K+4)}$, and $\boldsymbol{\mu} \in \mathbb{R}^{2V}$ is the mean shape. Any arbitrary shape sample $\mathbf{s}$ can now be represented in this model as $\hat{\mathbf{s}} = \boldsymbol{\mu} + \mathbf{\Phi}\mathbf{p}$, where

$$\mathbf{p} = \mathbf{\Phi}^T(\mathbf{s} - \boldsymbol{\mu}), \quad \mathbf{p} = (p_1, p_2, ..., p_{K+4})^T \in \mathbb{R}^{K+4},$$

represents the shape parameters of $\mathbf{s}$.

**Appearance Model:** To train the appearance model, each training image $\mathbf{I}_i$ is first warped from its shape $\mathbf{s}_i$ to the mean shape $\boldsymbol{\mu}$, and then vectorized to form $\mathbf{a}_i \in \mathbb{R}^L$. The appearance model is built by applying PCA to the matrix $\mathbf{A} = (\mathbf{a}_1, \mathbf{a}_2, ..., \mathbf{a}_N) \in \mathbb{R}^{L \times N}$, thus yields the mean appearance $\boldsymbol{\nu} \in \mathbb{R}^L$, the orthonormal appearance eigenvectors $\mathbf{\Psi} \in \mathbb{R}^{L \times M}$ (where $M < N$), and corresponding eigenvalues $\boldsymbol{\kappa} \in \mathbb{R}^M$. Any vectorized appearance sample $\mathbf{a}$ can then be represented by a set of appearance parameters $\mathbf{q}$, where $\mathbf{q} = (q_1, q_2, ..., q_M)^T \in \mathbb{R}^M$.

A trained AAM model is specified by its components: $\mathcal{A} = (\boldsymbol{\mu}, \mathbf{\Phi}, \boldsymbol{\lambda}, \boldsymbol{\nu}, \mathbf{\Psi}, \boldsymbol{\kappa})$. Fitting an image $\mathbf{I}$ to model $\mathcal{A}$ involves finding the best combination of shape and appearance parameters such that there exists a warp mapping the reconstructed shape into the reference frame; and the squared difference between the input image (subjected to the same warp) and the reconstructed image (based on the appearance parameters) is minimized.

### 2.2. Transfer Learning for AAMs

Although AAMs have seen tremendous success for a number of computer vision applications, their generalizablity is still sometimes challenging [10]. Specifically, when fitting is required on a very different set of images than those used to train the model, the fitting performance declines [10, 11]. This has led to the collection and annotation of many datasets (*e.g.* [8, 9]). If we have to fit AAMs to images that capture a different set of variations in expression, pose, illumination, or identity, ideally one would prefer to avoid the annotation step altogether or annotate only a handful of examples.

When a model is learned using only a few annotated examples, often it is not sufficiently expressive due to lack of enough variation [11]. What is the best way to make use of the already annotated datasets? How best to fuse the knowledge learned from multiple data sources and generalize it to new data? These questions warrant investigating the use of transfer learning for AAMs, wherein the goal is to transfer knowledge gained from previously available data (referred to as the *source* domain) to a new set of data (referred to as the *target* domain).

Transfer learning comes in several settings depending on whether the data are labeled in the source/target domain and whether the learning tasks in the source and target domains are the same [13]. In the context of AAM transfer learning, we focus on the situation when very few annotated examples are available for the target (T) domain, while the source (S) domain has a significant number of annotated examples, possibly coming from different datasets.

Let the model $\mathcal{A}_T = (\boldsymbol{\mu}_T, \boldsymbol{\Phi}_T, \boldsymbol{\lambda}_T, \boldsymbol{\nu}_T, \boldsymbol{\Psi}_{\mathbf{T}}, \boldsymbol{\kappa}_T)$ be learned from only a few target samples (*e.g.* $< 10$). Such a model leads to extreme overfitting, due to the lack of variation in the training set [14]. This effect becomes pronounced especially when the model is trained with high dimensional images [11], as is often the case. On the other hand, a model $\mathcal{A}_S = (\boldsymbol{\mu}_S, \boldsymbol{\Phi}_S, \boldsymbol{\lambda}_S, \boldsymbol{\nu}_S, \boldsymbol{\Psi}_S, \boldsymbol{\kappa}_S)$ trained only with source samples has significant variation, but a part of this variation might not be representative of the target domain. This can act as a confounding factor for the fitting process. There is thus a need to make use of however little knowledge is available from the target data, while simultaneously capitalizing on the large body of information available in the source domain. The goal of AAM transfer learning is to use both the target and the source samples to train a model $\mathcal{A}^* = (\boldsymbol{\mu}^*, \boldsymbol{\Phi}^*, \boldsymbol{\lambda}^*, \boldsymbol{\nu}^*, \boldsymbol{\Psi}^*, \boldsymbol{\kappa}^*)$, such that $\mathcal{A}^*$ outperforms both $\mathcal{A}_T$ and $\mathcal{A}_S$ on test samples from the target domain.

### 2.3. Baselines

The source-only and target-only models ($\mathcal{A}_S$ and $\mathcal{A}_T$) serve as a bare minimum baseline that a transfer learning approach must outperform. A useful transfer learning method should also outperform a model trained directly on the union of the source and target data denoted by $\mathcal{A}_{SUT} = (\boldsymbol{\mu}_{SUT}, \boldsymbol{\Phi}_{SUT}, \boldsymbol{\lambda}_{SUT}, \boldsymbol{\nu}_{SUT}, \boldsymbol{\Psi}_{SUT}, \boldsymbol{\kappa}_{SUT})$. The fourth baseline we consider is the subspace transfer (ST) method [11], which employs the mean shape and mean appearance of target model, *i.e.* $\boldsymbol{\mu}_{ST} = \boldsymbol{\mu}_T$ and $\boldsymbol{\nu}_{ST} = \boldsymbol{\nu}_T$. However, eigenvectors are evaluated by concatenating the source and target basis vectors, followed by a QR decomposition on matrices $[\boldsymbol{\Phi}_T, \boldsymbol{\Phi}_S]$ and $[\boldsymbol{\Psi}_T, \boldsymbol{\Psi}_S]$, to orthogonalize the basis vectors. Finally, the current state-of-the-art AAM transfer learning employs an instance-weighted (IW) approach by assigning weights to source domain samples based on importance sampling [11].

## 3. Subspace Selection

Our method[1] aims to transfer the knowledge gained from the source domain by selecting a subspace from the source dataset. First, we retain the target basis vectors $\boldsymbol{\Phi}_T$ as is, so that the information gained from target samples is not lost. Moreover, since the number of target examples assumed to be small, we propose to augment the target basis vectors with additional principal directions from the source data that are most representative of target space.

What should be our criterion to select source eigenvectors that are representative of the target subspace? Since all the basis vectors in $\boldsymbol{\Phi}_S$ and $\boldsymbol{\Phi}_T$ are of unit norm, every vector in $\boldsymbol{\Phi}_S$ is related to those in $\boldsymbol{\Phi}_T$ through a rotation, *i.e.*

$$\phi_{T_j} = \mathbf{R}_{i \to j} \phi_{S_i} \qquad \begin{cases} i \in \{1, ..., N_S\} \\ j \in \{1, ..., N_T\} \end{cases} \tag{1}$$

where $\phi_{S_i}$ and $\phi_{T_j}$ are the $i$-th and $j$-th eigenvectors in $\boldsymbol{\Phi}_S$ and $\boldsymbol{\Phi}_T$ respectively, and $\mathbf{R}_{i \to j}$ is the rotation between the two vectors. This makes the squared cosine of the angle between $\phi_{S_i}$ and $\phi_{T_j}$ as a natural measure of similarity between the two eigenvectors.

One measure of the overall similarity of the $\phi_{S_i}$ to the target subspace $\boldsymbol{\Phi}_T$, denoted $\gamma_i$, is defined as:

$$\gamma_i = \sum_{j=1}^{N_T} cos^2 \theta_{i \to j} \tag{2}$$

where $\theta_{i \to j}$ is the angle between $\phi_{S_i}$ and $\phi_{T_j}$. However, all the directions in $\boldsymbol{\Phi}_T$ are not equally important and, therefore, we propose that the similarity metric between $\phi_{S_i}$ and $\boldsymbol{\Phi}_T$ should also encode the relative importance of eigenvectors in $\boldsymbol{\Phi}_T$. Specifically, we define a weighted combination of individual similarities ($cos^2 \theta_{i \to j}$) such that each term is weighted by the corresponding eigenvalue associated with $\phi_{T_j}$, *i.e.*

[1] Our code is available at `https://github.com/azinasg/AAM_TL`.

$$\gamma_{i(\text{weighted})} = \sum_{j=1}^{N_T} \lambda_{T_j} cos^2\theta_{i\to j} \tag{3}$$

Because all eigenvectors have unit norm, we can write the above as follows:

$$\gamma_{i(\text{weighted})} = \boldsymbol{\phi}_{S_i}^T(\boldsymbol{\Phi}_T \boldsymbol{\Lambda}_T \boldsymbol{\Phi}_T^T)\boldsymbol{\phi}_{S_i} \tag{4}$$

where $\boldsymbol{\Lambda}_T$ is a diagonal matrix containing target eigenvalues along the diagonal. We note that the term in the parentheses is the eigen decomposition of the target covariance matrix, therefore

$$\gamma_{i(\text{weighted})} = \boldsymbol{\phi}_{S_i}^T(\frac{1}{N_T - 1}\mathbf{Z}_T\mathbf{Z}_T^T)\boldsymbol{\phi}_{S_i} \tag{5}$$

where $\mathbf{Z}_T \in \mathbb{R}^{2V\times N_T}$ contains the aligned and mean centered shape samples from the target.

We have thus shown that the weighted similarity metric between $\boldsymbol{\phi}_{S_i}$ and the target subspace $\boldsymbol{\Phi}_T$ is exactly equal to the variance of the target data in the direction of the $i$-th source eigenvector. In particular, we define the variance of the target data given source model as

$$\boldsymbol{\sigma}_{T|\boldsymbol{\Phi}_S}^2 = (\sigma_1^2, \sigma_2^2, ..., \sigma_{K_S}^2), \text{with } \ \sigma_i^2 = \frac{1}{N_T - 1}\boldsymbol{\phi}_{S_i}^T\mathbf{Z}_T\mathbf{Z}_T^T\boldsymbol{\phi}_{S_i} \tag{6}$$

So $\sigma_i^2$ is a measure of similarity between the $i$-th source eigenvector and target subspace. Therefore, in order to choose a subset of source eigenvectors that best represents the target subspace, we propose to sort the source eigenvectors in descending order of $\sigma_i^2$ (instead of the default ordering based on source eigenvalues,).

After rearranging the source eigenvectors, we pick up the top $D$ eigenvectors from the rearranged version of $\boldsymbol{\Phi}_S$ to get $\boldsymbol{\Phi}_D$. We treat $D$ as a hyper-parameter, which we determine by cross validation. We then concatenate the target basis vectors *i.e.* $\boldsymbol{\Phi}_T$ with $\boldsymbol{\Phi}_D$, and orthonormalize them using QR factorization to get $\boldsymbol{\Phi}^*$ to define our model. The steps involved in our subspace selection method are summarized in Algorithm 1.

## 3.1. Appearance

To evaluate the appearance part of our model, there are two differences in the procedure. First, since $\boldsymbol{\Psi}_S$ and $\boldsymbol{\Psi}_T$ are defined within two different base meshes *i.e.* $\boldsymbol{\mu}_S$ and $\boldsymbol{\mu}_T$, for finding projected target variance in appearance, we warp all eigenvectors of $\boldsymbol{\Psi}_S$ from given mean $\boldsymbol{\mu}_S$ to $\boldsymbol{\mu}_T$ with a piecewise affine warp $\mathcal{W}_{\boldsymbol{\mu}_S\to\boldsymbol{\mu}_T}(\boldsymbol{\Phi}_S)$. Second, after finding $\boldsymbol{\Psi}_D$, before QR decomposition, all appearance eigenvectors must be defined in the mean shape of the model, which again can be done easily by defining a piecewise affine warp.

**Algorithm 1** Subspace Selection Algorithm

**procedure** SELECTION($\boldsymbol{\Phi}_T, \boldsymbol{\Phi}_S, \mathbf{Z}_T$)
  **for** each $\boldsymbol{\Phi}_{S_i} \in \boldsymbol{\Phi}_S$ **do**
    $\sigma_i^2 = \frac{1}{N_T-1}\boldsymbol{\Phi}_{S_i}^T\mathbf{Z}_T\mathbf{Z}_T^T\boldsymbol{\Phi}_{S_i}$
  **end for**
  $\boldsymbol{\Phi}_D \leftarrow$ First $D$ basis vectors of $\boldsymbol{\Phi}_S$ arranged based on $\boldsymbol{\sigma}_{T|\boldsymbol{\Phi}_S}^2 = (\sigma_1^2, \sigma_2^2, ..., \sigma_{K_S}^2)$
  $\boldsymbol{\Phi}^* \leftarrow$ QR factorization($[\boldsymbol{\Phi}_T, \boldsymbol{\Phi}_D]$)
  **return** $\boldsymbol{\Phi}^*$
**end procedure**

# 4. Experimental Setup

## 4.1. Data

To compare our approach with other methods, we have conducted extensive experiments over six publicly available face datasets. In all, 555 samples were randomly selected from the following databases: LFPW [4], Helen [15], CK+ [9], iBUG [16], and AFW [17]. Images from the above datasets have ground truth annotations for 68 points. Images in the CK+ dataset cover seven posed expressions. The rest of the datasets cover a large variation in pose and illumination, and the majority of images are from young people and children with happy or neutral expressions.

In addition, we selected 320 examples from the UNBC-McMaster Shoulder Pain Expression Archive [8] by temporal downsampling (1 in 100) of videos in the dataset. The UNBC-McMaster dataset contains real expressions of pain from persons with shoulder injury. Each image in this data has ground truth annotation for 66 points. For other datasets which had 68 point annotations, the two additional points (inner corners of mouth) were removed so that annotations were consistent across all datasets. All experiments were thus done with 66 point annotations. In terms of the choice for source and target domains we considered the following two settings:

**Setting 1: Real expressions in target:** In this setting, UNBC-McMaster dataset was considered as the target domain, while the rest of the datasets were considered as the source domain. Five examples were randomly selected from the target domain to define the target training set. The test set consisted of 210 examples by excluding all the images from persons which were part of the training set. Since UNBC-McMaster data consists of real expressions of pain, in this setting the target domain consists of real expressions not present in the source.

**Setting 2: Posed expressions in target:** In this setting, CK+ dataset was considered as the target domain. The purpose of this setting is to present a further challenge to transfer learning methods by considering a target domain with multiple posed or fake expressions (*e.g.* sadness, anger,

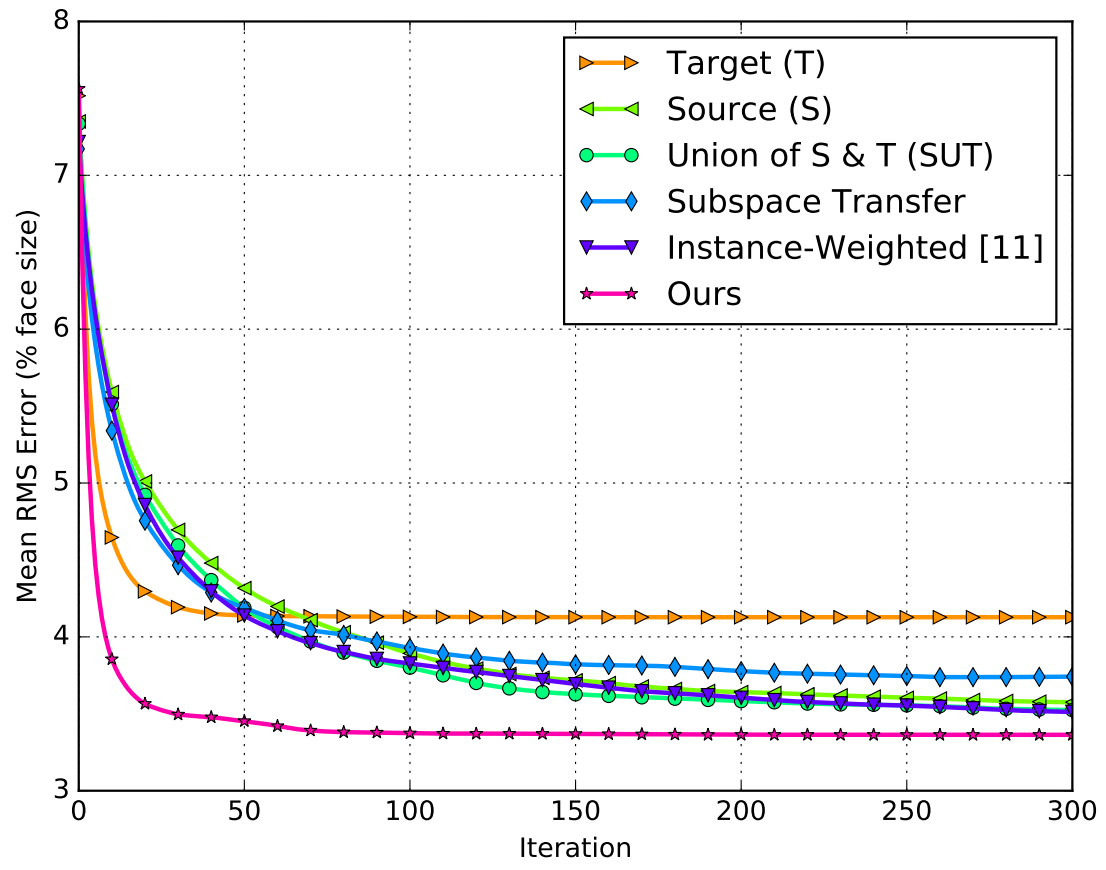

(a) RMS fitting error (Target: UNBC-McMaster)

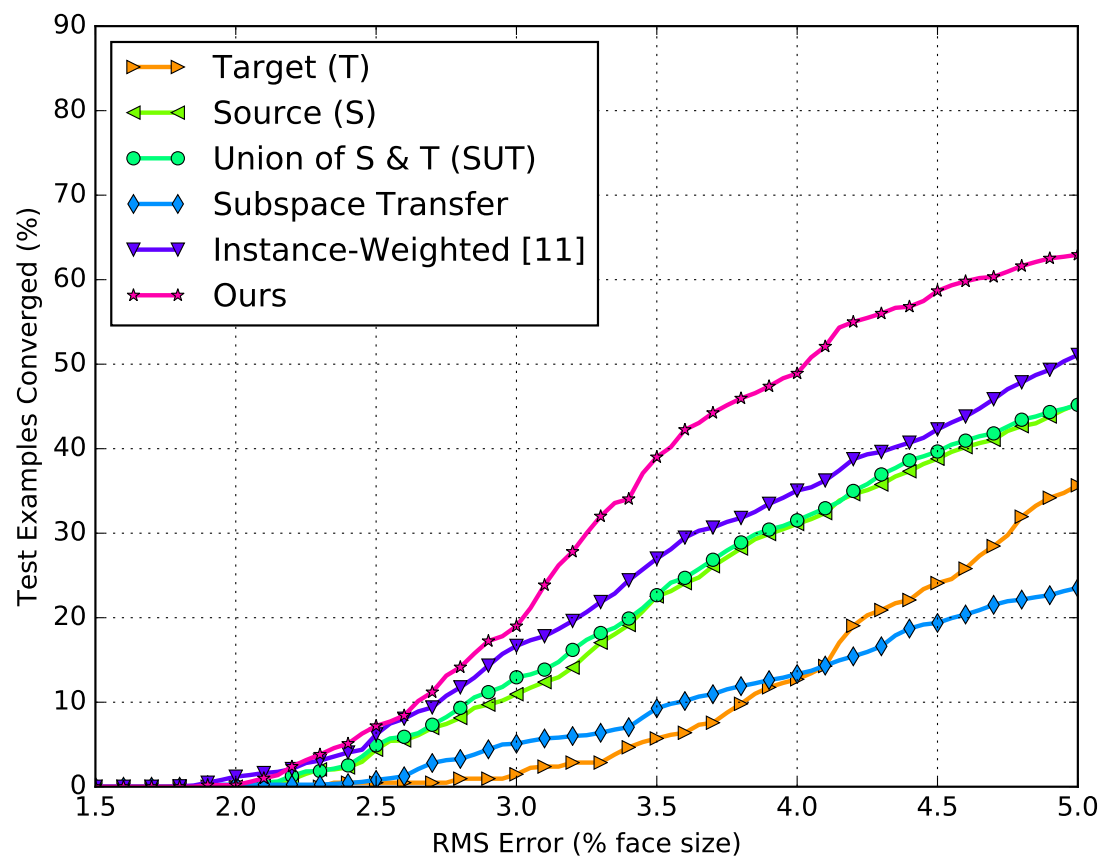

(b) % of test examples converged (Target: UNBC-McMaster)

Figure 2. Comparison of RMS fitting error and the percentage of converged test examples in Setting 1.

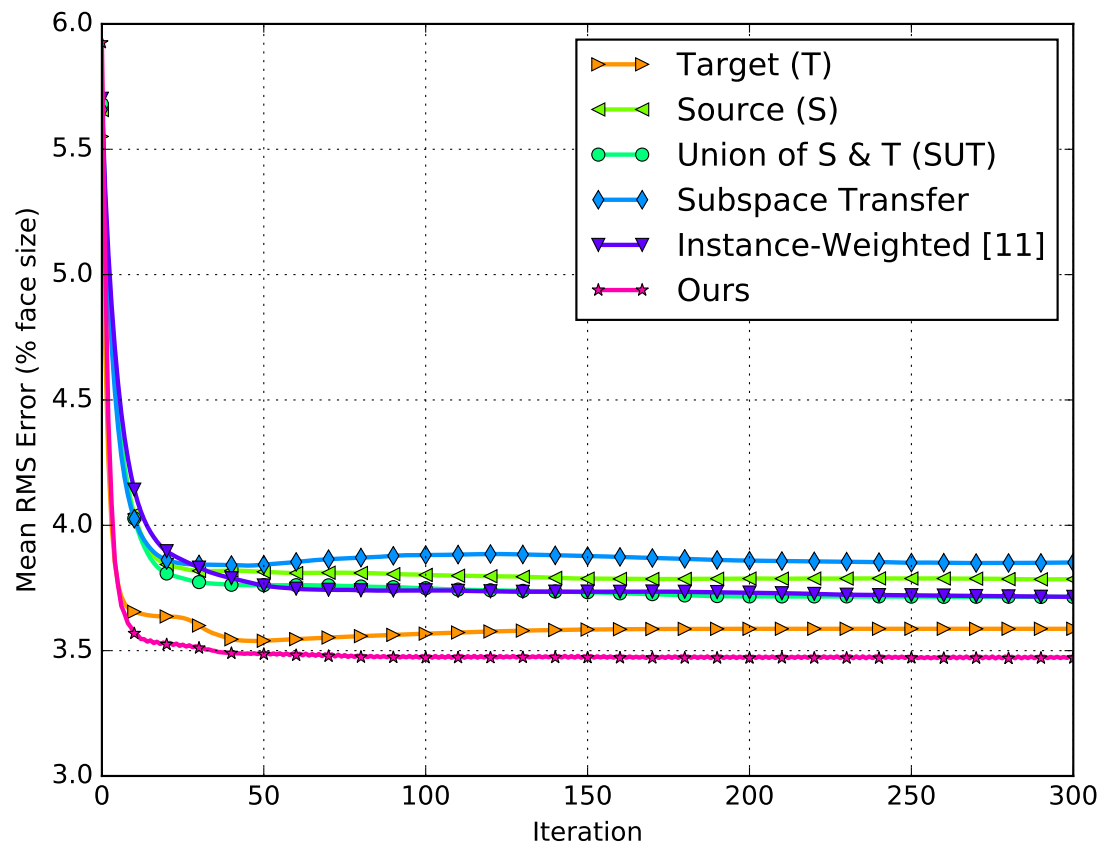

(a) RMS fitting error (Target: CK+)

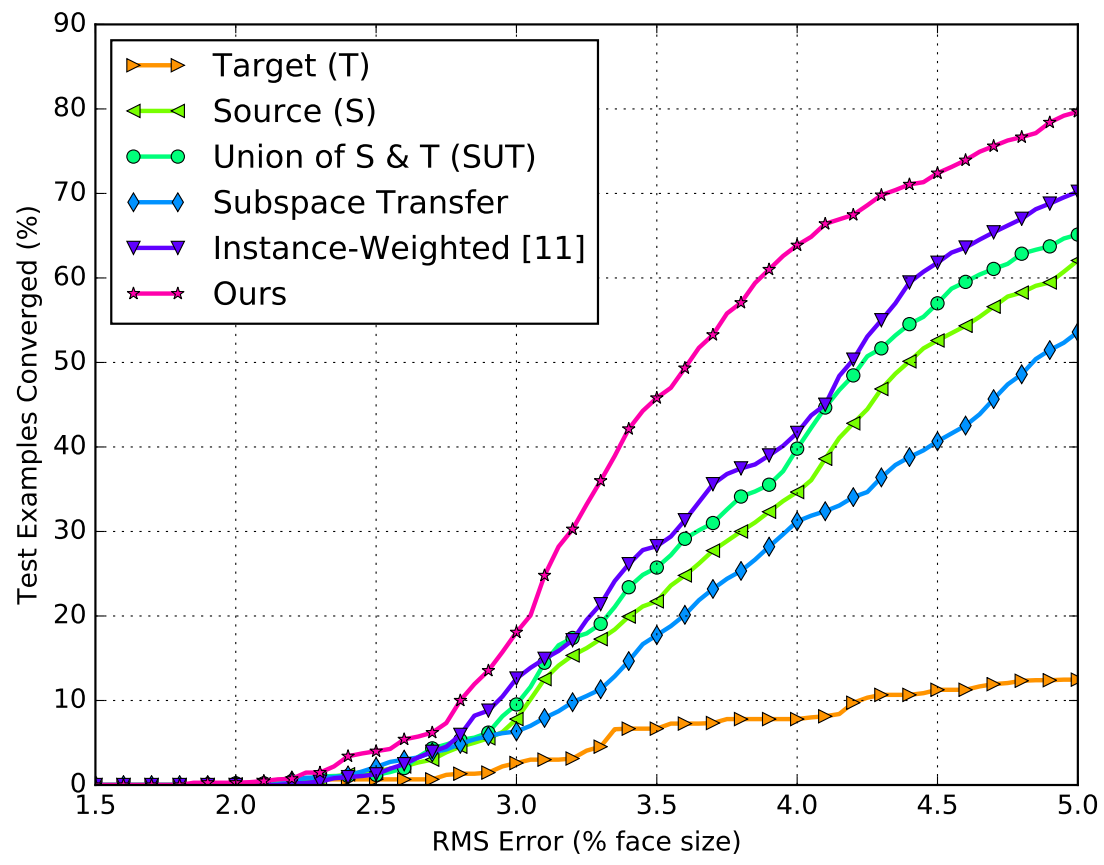

(b) % of test examples converged (Target: CK+)

Figure 3. Comparison of RMS fitting error and the percentage of converged test examples in Setting 2.

etc.) which are absent or substantially underrepresented in the source domain. For this setting, the source domain not only excluded the CK+ dataset (because it is the target domain), but also excluded UNBC-McMaster (which might include expression variations similar to CK+) to make the setting more challenging. For training, five examples were randomly picked from the target domain. The test set consisted of 150 examples by excluding all the images from persons which were part of the training set.

## 4.2. Fitting Details

For fitting, the Wiberg Inverse Compositional algorithm was used [2, 3]. We consider the fitting procedure as converged when the relative change in the cost function is very small ($< 10^{-5}$). The maximum number of iterations was set to 300. To initialize the fitting procedure, a bounding box is first fit around the face using the Viola-Jones face detector [18]. Then the mean shape of target ($\boldsymbol{\mu}_T$) is fit to the face bounding box by estimating a transformation (including only scale and translation). We call this initialization the base initialization. To avoid getting stuck in poor local minima we try 10 different perturbations around the base initialization by adding Gaussian noise in scale, translation, and rotation.

## 4.3. Performance Metrics

We use two standard criteria defined previously in the literature [3, 10] for evaluating AAM performance. The first criterion is the fitting accuracy. To quantify fitting accuracy, we measure the RMS error between the points of fitted shape and the ground truth landmark points normalized by the face size (average height and width of face) as suggested in [17]. The second criterion is the percentage of test examples that converge to the ground truth shape given a tolerance in the RMS fitting error (here, $10^{-5}$). Specifically, we analyze the percentage of test examples that converged to the ground truth as a function of the RMS error tolerance.

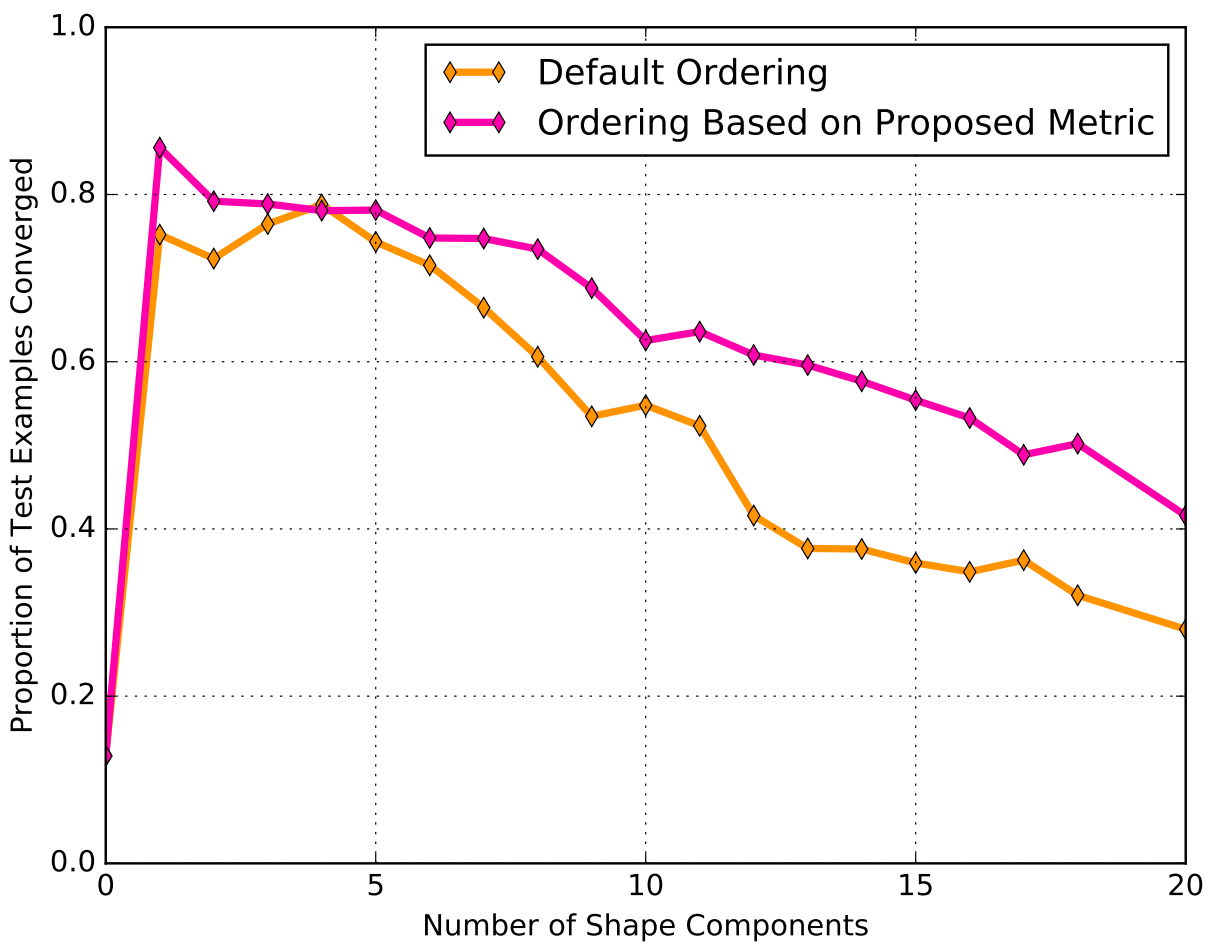

Figure 4. Percentage of converged test examples (with normalized RMS error less than 0.05) shown as a function of $D$, where $D$ represents the top $D$ source eigenvectors as ordered according to our proposed metric compared against the default ordering

## 5. Results

We compare different models in terms of the RMS error and percentage convergence for Setting 1 (UNBC-McMaster as target) in Figure 2. The curves in Figure 2(a) show the RMS error (averaged over examples for which the converged RMS error was less than 5% of the face size) as a function of iterations. The plots in Figure 2(b) show the percentage of test examples converged to the ground truth as a function of RMS tolerance in pixels. The corresponding curves for Setting 2 (CK+ as target) are shown in Figure 3. For both settings our approach outperforms all other methods in terms of RMS error as well as the percentage of test examples that converge to the ground truth. For our approach, the number of source principal components that were selected (*i.e.* the hyper-parameter $D$) was determined to be 3 for shape, and 30 for appearance using cross validation. Cross validation was performed by varying $D$ and picking the top $D$ source eigenvectors ordered according to their ability to capture variance in target data as determined in Equation 6.

In Figure 3, curves showing the two performance metrics should be interpreted together. For instance, in Figure 3(a) and 3(b), we see that the "Target" model has a good fitting accuracy over converged trials, while the percentage of convergence is very low. This is possibly due to the the lack of expressiveness of the model based only on target examples. On the other hand, the "Source" model has a higher rate of convergence, but the fitting accuracy is low. Also the "SUT" model performs well above the "Target" model and close to the "Source" model with a slight improvement resulting from the inclusion of target samples. The IW approach [11] has a small improvement in percentage of converged examples over previous models; but unexpectedly performs worse than the "Target" model in terms of the fitting accuracy, perhaps due to the source weight heuristics thereby affecting the target principal directions as well. Our approach improves the percentage of converged examples, and the fitting error is significantly decreased.

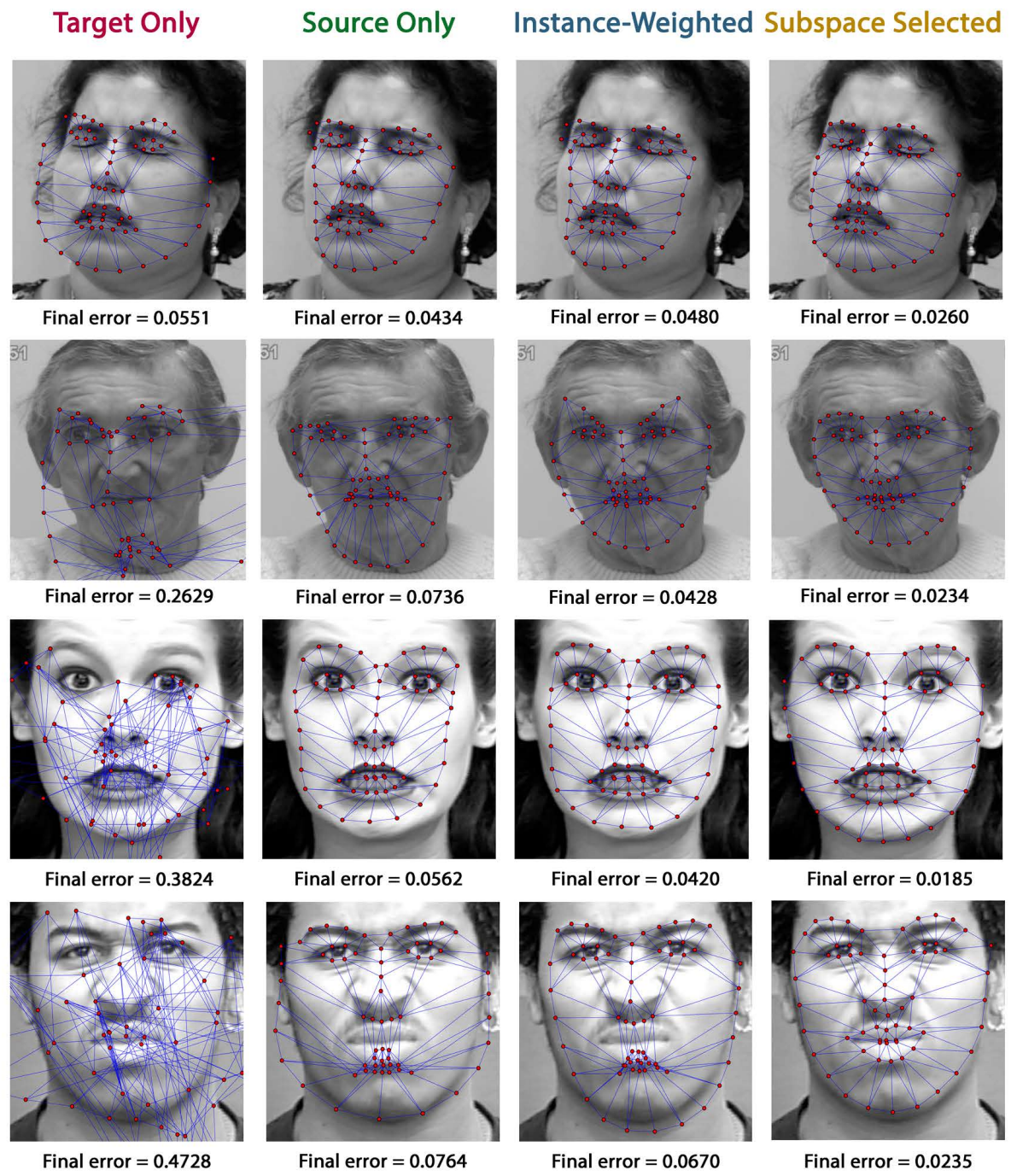

Figure 5. Representative examples for AAM fitting. The first two rows and the last two rows show RMS fitting error and a visualization of AAM fitting on two test examples from UNBC-McMaster dataset and CK+ respectively.

Figure 4 shows the percentage of converged trials with RMS error less than 0.05, obtained by picking top $D$ source eigenvectors as ordered by our metric (magenta plot). For comparison we also show the same by picking top $D$ eigenvectors based on the default ordering using their corresponding eigenvalues *i.e.* according to variance of source data. It shows that our approach outperforms this default ordering by $\sim$12 percentage points in convergence rate on average.

In Figure 5 we show the visual comparison of AAM fitting for different approaches. The first two rows show test samples from UNBC-McMaster dataset which include examples of pain expression and/or older adults not in the source domain of Setting 1. The last two rows show test samples from the CK+ dataset showing substantial fake expressions which were absent in the source domain of Setting 2. In all cases, the fitting results of our algorithm (last column) are closer to the actual landmark points and the RMS fitting error is also the minimum.

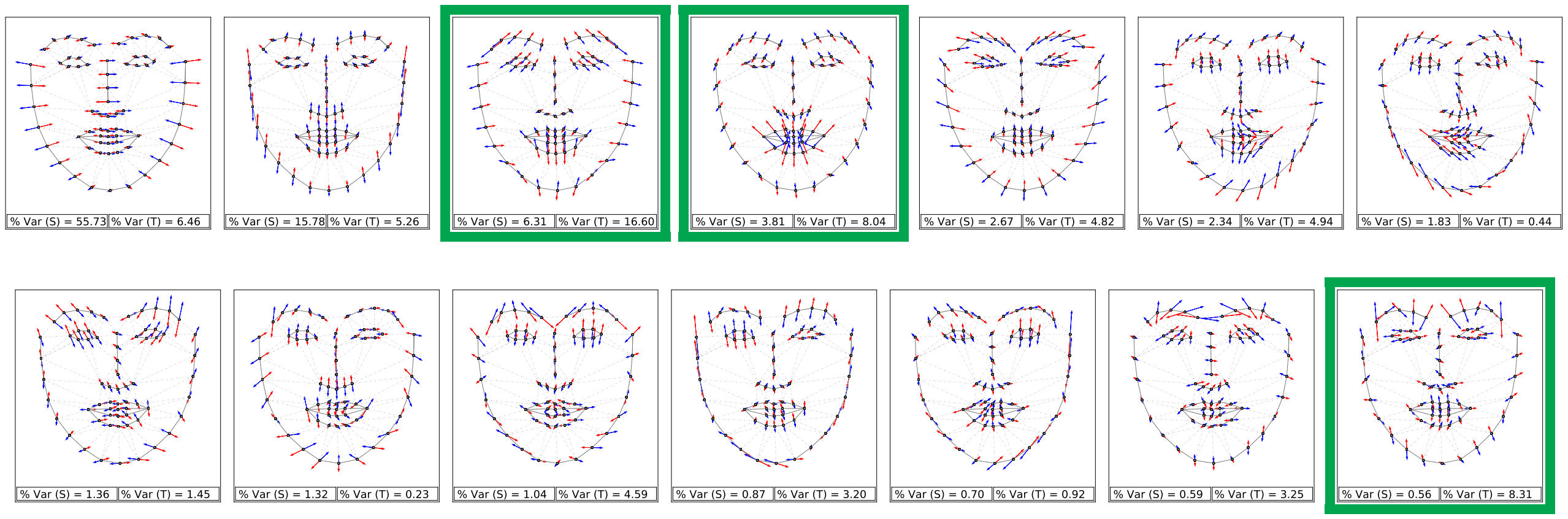

Figure 6. Visualization of Source principal directions. The three selected shape eigenvectors augmented to the target space are shown in green box. For every principal direction, the percentage variance captured in the source and target domains is also shown.

### 5.1. Analysis of Selected Source Directions

In this section we analyze the source principal directions picked by our subspace selection method. A visualization of source shape principal components for Setting 2 is shown in Figure 6. The arrows on the landmark points indicate the difference vector between the eigenvector and the mean shape. The three selected eigenvectors are enclosed in green boxes. For every principal component, we also show the percentage variance of the source and target samples when projected onto the component. The first two selected eigenvectors are principal directions which cover sizable variance around the mouth and eye region, explaining why they were selected because the target data set has significant expression variation around these regions. Similarly, the third selected eigenvector has dominant motion around the eyebrow region.

For further analysis, we looked into the source examples that are best explained by the selected directions (Figure 7). As can be seen, these are the source examples which are explained well by direction vectors capturing target variance and show more vivid expressions around the mouth and eye regions. On top of these images in Figure 7, we also show the weight assigned to them by the instance-weighted approach. As a comparison, in the last row we show source examples which were highly weighted by the same approach. We see that neutral and smiling faces which are in majority in the source get weighted by the heuristic IW approach, while the source examples which are perhaps more representative of the target data are given low weights. However, by selecting only those eigenvectors which capture target variance, we see that we are able to encode the information from just the right source examples, and transfer it to the target domain.

## 6. Conclusions and Future Work

In this paper we have presented a transfer learning method for AAMs. Our method is based on selecting a subset of source eigenvectors that are most representative of the target subspace. This selection is based on a metric that captures the directional similarity between source eigenvectors and the target subspace. We have shown an equivalence between the similarity metric and the variance captured by source eigenvectors in the target space – which was our basis for selecting a subset of source principal components. We have conducted our experiments over six publicly available datasets and have tested challenging scenarios wherein the variations in the target domain were substantially different from those in the source domain. Our method outperforms the state of the art AAM transfer learning approach [11] and other baselines. We note that the experimental setups tested in [11] were significantly less challenging as examples for source and target domains came from the same dataset.

We have demonstrated that even when only a handful of annotated target examples are available, superior AAM fitting could be achieved using our transfer learning approach. Collecting and annotating datasets is a painstaking and time consuming step. This often becomes an obstacle in the way of using AAM fitting in novel application settings. Our work can have potential impact on extending the use of AAMs to new application domains. For instance, within clinical settings, there is growing interest to use facial expression analysis for healthcare applications such as pain monitoring in older adults, assessing signs of depression, aggression, and agitation. In contexts such as the above, collection of large datasets is doubly challenging because of patient confidentiality and privacy issues. In the future we plan to investigate how well our method generalizes to applications not involving face analysis such as brain seg-mentation.

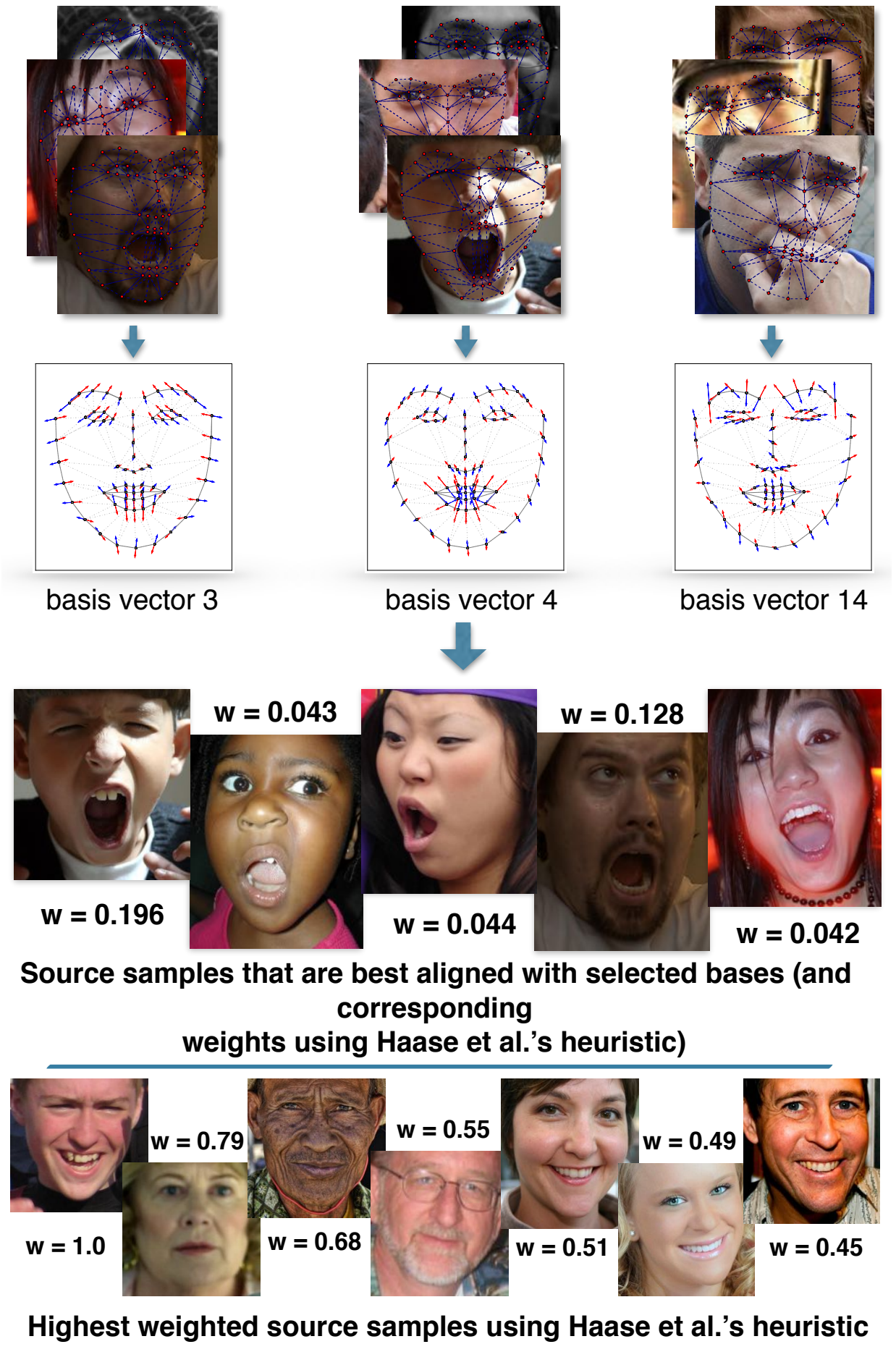

Figure 7. Source examples that are best explained by the selected eigenvectors on the basis of target variance. Within the source data these examples turn out to be less frequent examples. We also show the weights assigned to these examples using the instance-weighted approach [11]. Last row shows source samples that were assigned highest weights based on [11].

## 7. Acknowledgments

This research was supported by the AGE-WELL NCE, the Canadian Institutes of Health Research (CIHR), and the Toronto Rehabilitation Institute University Health Network (TRI-UHN).

## References

[1] Timothy F Cootes, Gareth J Edwards, Christopher J Taylor, et al. Active appearance models. *IEEE Transactions on PAMI*, 23(6):681–685, 2001.

[2] Joan Alabort-i Medina, Epameinondas Antonakos, James Booth, Patrick Snape, and Stefanos Zafeiriou. Menpo: A comprehensive platform for parametric image alignment and visual deformable models. In *ACM International Conference on Multimedia*, pages 679–682. ACM, 2014.

[3] Joan Alabort-i Medina and Stefanos Zafeiriou. A unified framework for compositional fitting of active appearance models. *IJCV*, pages 1–39, 2016.

[4] Peter N Belhumeur, David W Jacobs, David J Kriegman, and Neeraj Kumar. Localizing parts of faces using a consensus of exemplars. *IEEE transactions on PAMI*, 35(12):2930–2940, 2013.

[5] Sumaira Kausar and M. Younus Javed. A survey on sign language recognition. In *Frontiers of Information Technology*, pages 95–98, Washington, DC, USA, 2011. IEEE Computer Society.

[6] T. Heap and D. Hogg. Towards 3d hand tracking using a deformable model. In *Automatic Face and Gesture Recognition*, pages 140–145, Oct 1996.

[7] Z.R. Luo, X.J. Zhuang, R.Z. Zhang, J.Q. Wang, C. Yue, and X. Huang. Automated 3d segmentation of hippocampus based on active appearance model of brain mr images for the early diagnosis of alzheimer's disease. *Minerva Med.*, 105:157–165, 2014.

[8] Patrick Lucey, Jeffrey F Cohn, Kenneth M Prkachin, Patricia E Solomon, and Iain Matthews. Painful data: The unbc-mcmaster shoulder pain expression archive database. In *Automatic Face and Gesture Recognition and Workshops*, pages 57–64. IEEE, 2011.

[9] Patrick Lucey, Jeffrey F Cohn, Takeo Kanade, Jason Saragih, Zara Ambadar, and Iain Matthews. The extended cohn-kanade dataset (ck+): A complete dataset for action unit and emotion-specified expression. In *CVPR Workshops*, pages 94–101. IEEE, 2010.

[10] Joan Alabort-i Medina and Stefanos Zafeiriou. Bayesian active appearance models. In *Proceedings of the IEEE CVPR*, pages 3438–3445, 2014.

[11] Daniel Haase, Erid Rodner, and Joachim Denzler. Instance-weighted transfer learning of active appearance models. In *CVPR*, pages 1426–1433, 2014.

[12] Iain Matthews and Simon Baker. Active appearance models revisited. *IJCV*, 60(2):135–164, 2004.

[13] Sinno Jialin Pan and Qiang Yang. A survey on transfer learning. *IEEE Transactions on knowledge and data engineering*, 22(10):1345–1359, 2010.

[14] Ralph Gross, Iain Matthews, and Simon Baker. Generic vs. person specific active appearance models. *Image and Vision Computing*, 23(12):1080–1093, 2005.

[15] Vuong Le, Jonathan Brandt, Zhe Lin, Lubomir Bourdev, and Thomas S Huang. Interactive facial feature localization. In *ECCV*, pages 679–692. Springer, 2012.

[16] Christos Sagonas, Georgios Tzimiropoulos, Stefanos Zafeiriou, and Maja Pantic. 300 faces in-the-wild challenge: The first facial landmark localization challenge. In *IEEE ICCV Workshops*, pages 397–403, 2013.

[17] Xiangxin Zhu and Deva Ramanan. Face detection, pose estimation, and landmark localization in the wild. In *CVPR*, pages 2879–2886. IEEE, 2012.

[18] Paul Viola and Michael J Jones. Robust real-time face detection. *IJCV*, 57(2):137–154, 2004.